\documentclass{article}

\PassOptionsToPackage{square,numbers}{natbib}
\usepackage[preprint]{neurips_2026}

\usepackage[utf8]{inputenc} 
\usepackage[T1]{fontenc}    
\usepackage{hyperref}       
\usepackage{url}            
\usepackage{amsfonts}       
\usepackage{amsmath}
\usepackage{amssymb}
\usepackage{nicefrac}       
\usepackage{microtype}      
\usepackage{xcolor}         
\usepackage{comment}
\usepackage{booktabs}
\usepackage{graphicx}
\usepackage{multirow}

\usepackage{wrapfig}
\usepackage{enumitem}
\usepackage{listings}
\usepackage[most]{tcolorbox}
\usepackage{fvextra}
\usepackage{caption} 
\usepackage{mdframed}
\usepackage{caption}
\usepackage{subcaption}
\usepackage[scaled]{helvet}
\definecolor{darkgraybar}{RGB}{90,90,90}
\definecolor{lightgraybg}{RGB}{245,245,245}

\newtcolorbox{promptbox}[1]{
    enhanced,
    width=0.95\linewidth,
    colback=lightgraybg,
    colframe=darkgraybar,
    coltitle=white,
    colbacktitle=darkgraybar,
    title=#1,   
    fonttitle=\bfseries,
    arc=2mm,
    boxrule=0.6pt,
    left=3mm,
    right=3mm,
    top=2mm,
    bottom=2mm,
    fontupper=\ttfamily\small
}

\newtcolorbox{orangebox}[1]{
    enhanced,
    colback=orange!3,
    colframe=orange,
    coltitle=white,
    colbacktitle=orange,
    title=#1,
    fonttitle=\small\bfseries,
    arc=4pt,
    boxrule=1.5pt,
    left=4pt, right=4pt, top=4pt, bottom=4pt,
    fontupper=\fontsize{6}{8}\selectfont
}

\newtcolorbox{bluebox}[1]{
    enhanced,
    colback=blue!3,
    colframe=blue!70,
    coltitle=white,
    colbacktitle=blue!70,
    title=#1,
    fonttitle=\small\bfseries,
    arc=4pt,
    boxrule=1.5pt,
    left=4pt, right=4pt, top=4pt, bottom=4pt,
    fontupper=\fontsize{6}{8}\selectfont
}

\newtcolorbox{greenbox}[1]{
    enhanced,
    colback=green!3,
    colframe=green!60!black,
    coltitle=white,
    colbacktitle=green!60!black,
    title=#1,
    fonttitle=\small\bfseries,
    arc=4pt,
    boxrule=1.5pt,
    left=4pt, right=4pt, top=4pt, bottom=4pt,
    fontupper=\ttfamily\fontsize{6}{8}\selectfont
}

\newtcolorbox{purplebox}[1]{
    enhanced,
    colback=violet!3,
    colframe=violet!70,
    coltitle=white,
    colbacktitle=violet!70,
    title=#1,
    fonttitle=\small\bfseries,
    arc=4pt,
    boxrule=1.5pt,
    left=4pt, right=4pt, top=4pt, bottom=4pt,
    fontupper=\fontsize{6}{8}\selectfont
}

\newtcolorbox{redbox}[1]{
    enhanced,
    colback=red!3,
    colframe=red!70,
    coltitle=white,
    colbacktitle=red!70,
    title=#1,
    fonttitle=\small\bfseries,
    arc=4pt,
    boxrule=1.5pt,
    left=4pt, right=4pt, top=4pt, bottom=4pt,
    fontupper=\fontsize{6}{8}\selectfont
}

\lstdefinestyle{rawtext}{
  basicstyle=\ttfamily\small,
  breaklines=true,
  breakatwhitespace=false,
  breakindent=0pt,
  frame=none,
  keepspaces=true,
  columns=flexible,
  lineskip=-0.5pt,
}

\title{Answer-then-Edit: Reasoning Skeleton Editing for Anti-Distillation with Preserved Utility}

\author{
Fan Li$^{1*}$, 
Mengting Pan$^{1}$\thanks{Equal contribution.}, 
Sijia Xu$^{1}$\footnotemark[1], 
Xiaoyang Wang$^{1}$, 
Chen Chen$^{2}$, 
Wenjie Zhang$^{1}$ \\
$^{1}$School of Computer Science and Engineering, UNSW Sydney \\
$^{2}$School of Artificial Intelligence, Shenzhen University \\
\texttt{\{fan.li, mengting.pan, sijia.xu, xiaoyang.wang1, wenjie.zhang\}@unsw.edu.au} \\
\texttt{chen\_chen@uow.edu.au}
}

\begin{document}

\maketitle

\begin{abstract}
Proprietary large language models (LLMs) entail substantial intellectual and financial investment, making them valuable intellectual property (IP). However, even when deployed via black-box APIs, these models remain vulnerable to unauthorized knowledge distillation, which allows adversaries to cheaply extract and replicate model capabilities. To address this issue, anti-distillation (AD) has been proposed to generate defensive outputs that hinder distillation effectiveness, overcoming the limitation of watermarking-based approaches that rely on post-hoc verification. However, existing AD methods based on internal model perturbations struggle to balance anti-distillability and utility (e.g., answer accuracy and naturalness) of reasoning traces, with stronger defenses often causing significant utility loss. To fill this gap, we propose \textbf{\underline{S}}keleton-\textbf{\underline{G}}uided \textbf{\underline{R}}easoning \textbf{\underline{E}}diting (SGRE), an \textit{Answer-then-Edit} framework that performs post-hoc trace modification for anti-distillation. In the answer stage, the teacher model first generates clean reasoning traces, preserving the original reasoning accuracy while enabling more flexible control over trace naturalness. In the editing stage, we draw inspiration from Cognitive Load Theory (CLT) and introduce a three-stage strategy consisting of reasoning skeleton extraction, skeleton graph coarsening, and skeleton verbalization. These operations jointly perturb reasoning structures and augment textual complexity to amplify extraneous load on student models, hindering their acquisition of underlying reasoning patterns.
Extensive experiments across diverse LLMs demonstrate that SGRE achieves state-of-the-art performance in reducing distillation effectiveness, while maintaining lossless reasoning accuracy and superior trace naturalness.
\end{abstract}

\section{Introduction}
Large language models (LLMs) have demonstrated remarkable reasoning capabilities and serve as foundation models to solve complex tasks across diverse domains~\cite{achiam2023gpt,liu2024deepseek,yang2025qwen3}. Developing such models requires substantial resources, including large-scale data curation, expensive computational budgets, specialized expertise, and significant human effort, making them valuable intellectual property (IP). To safeguard these assets, commercial model providers typically restrict access to model internals and expose them solely through APIs. 
However, recent studies show that reasoning traces generated by proprietary LLMs can be exploited for knowledge distillation (KD)~\cite{ho2023large,grattafiori2024llama,guo2025deepseek,lim2026graykd}, transferring knowledge from large, powerful teacher models to smaller, more efficient student models. This enables competitors to cheaply replicate frontier model performance, posing significant IP risks.

To defend against this threat, LLM watermarking~\cite{he2022protecting,kirchenbauer2023watermark} and fingerprinting~\cite{xu2024instructional,ye2026securing} have been proposed by embedding identifiable signals into query responses, allowing the detection of misuse or unauthorized replication in distilled models. However, these strategies primarily offer post-hoc verification and fail to prevent unauthorized capability extraction through knowledge distillation. 
\begin{wrapfigure}{r}{0.55\linewidth}
\centering
\includegraphics[width=1\linewidth]{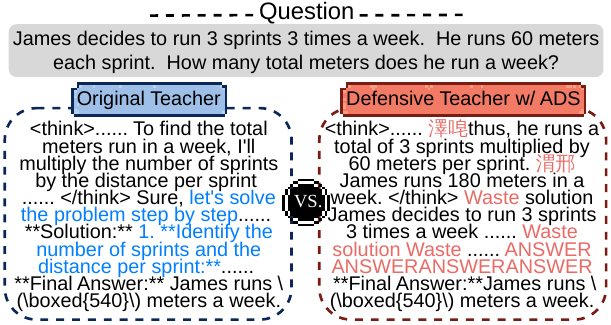}
\caption{Naturalness of anti-distilled traces.}
\vspace{-10pt}
\label{fig:1}
\end{wrapfigure}
To fill this gap, anti-distillation (AD)~\cite{savaniantidistillation,liDOGe} has attracted increasing attention as a technique for generating defensive reasoning traces that hinder effective distillation. 
Antidistillation Sampling (ADS)~\cite{savaniantidistillation} pioneers AD by perturbing the teacher’s next-token distribution with a gradient-based penalty, which increases the loss of student models trained on the generated traces. DOGe~\cite{liDOGe} involves adversarially fine-tuning the teacher’s final linear layer to generate outputs that mislead distillation.

Despite their effectiveness, existing AD methods struggle to achieve a desirable trade-off between the utility (e.g., teacher answer accuracy and trace naturalness) and the anti-distillability of generated reasoning traces. This limitation is reflected in two key aspects: (i) significantly reducing the distillation effectiveness of reasoning traces often comes at the cost of notable drops in answer accuracy. For example, as shown in Table~\ref{tab:main_results}, on GSM8K, ADS reduces the distillation performance of Llama-3.2-3B by 37.88\%, but incurs over a 19.03\% drop in reasoning accuracy. In contrast, DOGe preserves the teacher’s reasoning ability but degrades student performance by only 4.38\%.
(ii) Improving the anti-distillability of reasoning traces often leads to a noticeable decline in language naturalness. As illustrated in Figure~\ref{fig:1}, compared with the clean outputs from the original teacher, ADS produces traces with irregular tokens, incoherent structure, and grammatical errors, which undermine their semantic interpretability and impair the user experience of commercial LLM services.

We analyze that these trade-off issues may stem from model-internal perturbation paradigms. Specifically:
(i) ADS directly perturbs the logit distribution during decoding, introducing uncontrollable sampling noise that makes semantic coherence difficult to maintain and consequently degrades both reasoning accuracy and naturalness; 
(ii) DOGe performs anti-distillation through adversarial optimization on model parameters, where designing effective adversarial loss can be challenging and often relies on specific assumptions~\cite{liDOGe}, potentially limiting its practical effectiveness.
These observations motivate us to propose an \textit{Answer-then-Edit} paradigm, where we perform post-hoc editing on teacher-generated traces rather than perturbing model internals. This design preserves the teacher’s original reasoning capability and answer accuracy, while enabling anti-distillation at the textual and semantic levels. Consequently, this framework offers better controllability over readability and naturalness while avoiding excessive decoding noise and unreliable optimization objectives.

To design an effective post-hoc trace editing strategy, we draw on Cognitive Load Theory (CLT)~\cite{sweller1988cognitive,sweller1994cognitive,paas2003cognitive}, a well-established cognitive theory widely used in educational research for instructional design.
CLT posits that human cognitive capacity is limited and learning effectiveness is jointly influenced by three types of cognitive load: intrinsic load, extraneous load, and germane load. Specifically, intrinsic load reflects the inherent complexity of a learning task and is typically fixed for a given problem. Extraneous load is induced by the way instructional materials and tasks are presented, while germane load corresponds to the cognitive resources devoted to the formation of underlying knowledge schemas.
The core principle of CLT is that effective instructional design should minimize unnecessary extraneous load induced by instructional presentation, allowing more cognitive capacity to be allocated to germane load for effective schema acquisition~\cite{sweller1998cognitive}. 
For example, instructional materials with sufficient and well-structured content, clear logical organization, and explicit language expression can reduce unnecessary information-processing burden and improve student learning~\cite{sweller1994cognitive}. 
This motivates us to increase presentation complexity in reasoning traces during anti-distillation, thereby diverting student models from the core reasoning process and preventing the formation of underlying reasoning patterns.


Based on this insight, we propose \textbf{\underline{S}}keleton-\textbf{\underline{G}}uided \textbf{\underline{R}}easoning \textbf{\underline{E}}diting (SGRE), a post-hoc anti-distillation method that edits reasoning traces in terms of both reasoning logic and textual expression. Specifically, SGRE consists of three stages: reasoning skeleton extraction, skeleton graph coarsening, and skeleton verbalization. First, we extract a reasoning skeleton graph from teacher-generated traces using a small LLM, summarizing key reasoning steps and their dependencies into a compact structure. This process compresses auxiliary details and obscures explicit logical connections, making the underlying reasoning process harder to reconstruct. Then, we perform structured coarsening on the skeleton graph, disrupting the granularity of reasoning steps and entangling the logical structure. Finally, we verbalize the coarsened skeleton into the output trace using the teacher model, injecting controlled textual complexity while preserving overall correctness and naturalness.
Our main contributions are as follows:
\begin{itemize}[leftmargin=15pt]
    \item We identify that existing anti-distillation methods based on internal perturbations to the teacher model fail to achieve a desirable trade-off between utility and anti-distillability, and introduce an \textit{Answer-then-Edit} paradigm that decouples anti-distillation from the teacher reasoning process.
    \item Inspired by Cognitive Load Theory (CLT), we propose Skeleton-Guided Reasoning Editing (SGRE), a novel post-hoc anti-distillation framework that combines reasoning structure perturbation and textual complexity augmentation to amplify extraneous load in reasoning traces, thereby hindering student learning of reasoning patterns.
    \item Extensive experiments across diverse LLM architectures demonstrate that SGRE consistently achieves state-of-the-art performance in reducing distillation effectiveness while maintaining reasoning accuracy. In addition, naturalness evaluation shows that SGRE effectively mitigates readability degradation during anti-distillation, highlighting its practical usability.
\end{itemize}

\label{intro}


\section{Preliminary}
\label{preliminary}

\noindent \textbf{SFT-based Knowledge Distillation.}
In this work, following~\cite{savaniantidistillation,liDOGe}, we focus on supervised fine-tuning (SFT)-based knowledge distillation, which is widely adopted in practical text-based KD settings. Specifically, let $\mathcal{T}$ denote a large teacher model and $\mathcal{S}$ denote a smaller student model. 
Given a set of queries $\mathcal{Q} = \{ q_i \}_{i=1}^{n}$, the teacher autoregressively generates a reasoning trace
$r_i = \mathcal{T}(q_i)= (y_{i,1}, \ldots, y_{i,L_i})$ 
for each query $q_i$ as:
\begin{equation}
P_{\mathcal{T}}(r_i \mid q_i)
=
\prod_{t=1}^{L_i}
P_{\mathcal{T}}(y_{i,t} \mid q_i, y_{i,<t}),
\end{equation}
where $y_{i,1}, \ldots, y_{i,L_i}$ denote the generated response tokens.
The resulting query-response pairs form a distillation dataset $D=\{(q_i, r_i)\}_{i=1}^{n}$, on which the student $\mathcal{S}$ is trained via the following objective:
\begin{equation} 
\min_{\theta_{\mathcal{S}}} - \mathbb{E}_{(q_i, r_i)\sim D} \left[ \sum_{t=1}^{L_i} \log P_{\mathcal{S}}\bigl(y_{i,t} \mid q_i, y_{i,<t}\bigr) \right],
\end{equation}
where $\mathcal{\theta_{S}}$ denotes the model parameters of $\mathcal{S}$.

\noindent \textbf{Problem Formulation.} Let $\mathcal{T}$ be the original teacher and $\mathcal{T}^\prime = \mathcal{A}(\mathcal{T})$ be the defensive teacher, where $\mathcal{A}$ denotes the anti-distillation algorithm.
Given a query set $\mathcal{Q} = \{q_i\}_{i=1}^n$, we construct the corresponding distillation datasets
$\mathcal{D}_{\text{train}}=
\{(q_i, \mathcal{T}(q_i))\}_{i=1}^n$
and
$
\mathcal{D}_{\text{train}}^\prime=\{(q_i, \mathcal{T}^\prime(q_i))\}_{i=1}^n
$, respectively.
Let $\mathcal{Q}_{\text{test}}$ be a test query set, and let 
$\mathrm{Perf}(\cdot; \mathcal{Q}_{\text{test}})$ denote the evaluation metric (e.g., answer accuracy) on $\mathcal{Q}_{\text{test}}$. 
Let $\mathcal{S}(\mathcal{D})$ denote the student model trained on dataset $\mathcal{D}$.
The goal of anti-distillation is to find an algorithm $\mathcal{A}$ that minimizes the reasoning performance of the student trained on $\mathcal{D}_{\text{train}}^\prime$, while preserving the teacher utility. Formally, the objective is:
\begin{equation}
\begin{aligned}
\min_{\mathcal{A}} \quad 
& \mathrm{Perf}\big(\mathcal{S}(\mathcal{D}_{\text{train}}^\prime); \mathcal{Q}_{\text{test}}\big) \\
\text{s.t.} \quad 
& \left| \mathrm{Perf}(\mathcal{T}^\prime; \mathcal{Q}_{\text{test}}) - \mathrm{Perf}(\mathcal{T}; \mathcal{Q}_{\text{test}}) \right| \le \epsilon,
\end{aligned}
\end{equation}
where $\epsilon$ is a small tolerance for performance degradation.

\section{Methodology}
\label{method}

\begin{figure}[!t]
\centering
\includegraphics[width=1\columnwidth]{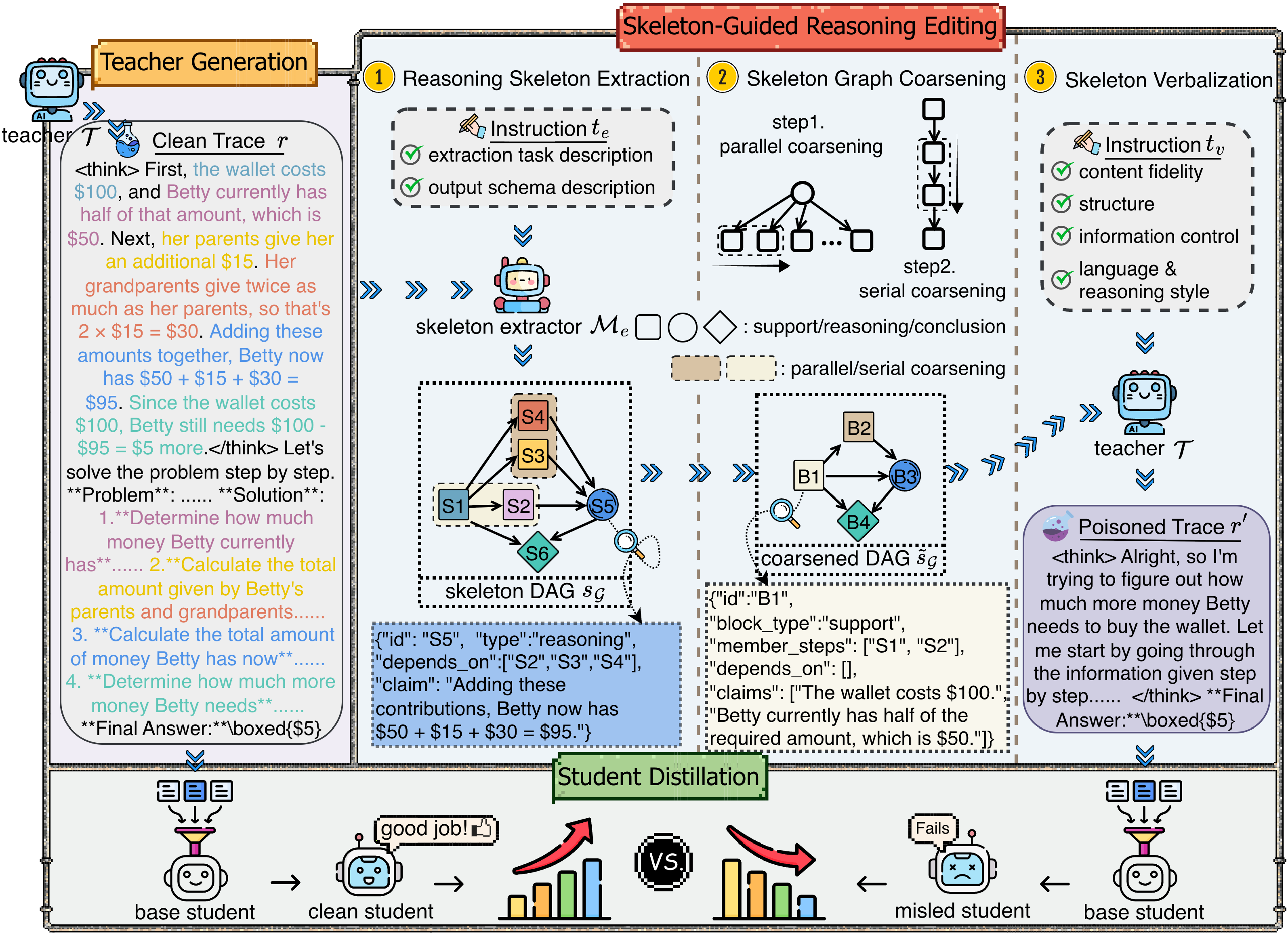} %
\caption{Overall framework of SGRE.}
\vspace{-1em}
\label{fig:framework}
\end{figure}

In this section, we propose SGRE, a novel post-hoc anti-distillation framework that consists of three essential modules. An overview of the proposed framework is illustrated in Figure~\ref{fig:framework}. The first module, termed reasoning skeleton extraction (RSE), converts the teacher-generated trace into a compact skeleton graph by compressing informative textual content and obscuring explicit logical dependencies. Then, skeleton graph coarsening (SGC) progressively merges reasoning steps in the skeleton graph to reduce the granularity of the reasoning process, making the underlying reasoning structure more difficult to reconstruct. Finally, skeleton verbalization (SV) transforms the coarsened skeleton graph into natural language with additional textual complexity. By combining reasoning structure perturbation and textual complexity augmentation, SGRE increases the information processing burden during distillation, thereby hindering student learning of underlying reasoning patterns.


\subsection{Reasoning Skeleton Extraction}

In human learning, overly abstracted and fragmented instructional materials often make information processing more difficult and impose additional cognitive burden on students, hindering the construction of clear reasoning schemas~\cite{sweller1994cognitive}. Motivated by this, reasoning skeleton extraction is designed to transform the teacher-generated trace $r$ by compressing both the content within reasoning steps and the logical connections between them. Such abstraction obscures the explicit step-by-step reasoning, thereby hindering student models from effectively learning the reasoning patterns during distillation.

To achieve this, we employ a small language model (SLM), denoted by $\mathcal{M}_e$, as the skeleton extractor for efficient reasoning abstraction. Given the teacher-generated reasoning trace $r$ and an extraction template $t_e$, the extractor produces a structured skeleton representation $s=\mathcal{M}_e(r,t_e)$. The prompt template for skeleton extraction is provided in Figure~\ref{fig:skeleton_prompt}. We describe the extraction process from two aspects: the \textit{Extraction Task Description} and the \textit{Output Schema}.

\noindent \textbf{Extraction Task Description.} 
The extractor is instructed to identify the underlying logical structure of the reasoning trace by focusing on abstract reasoning units and their dependencies. Surface-level phrasing and procedural details are removed, while the final answer is preserved. The resulting skeleton is required to remain human-interpretable but not directly usable as a procedural solution.

\noindent \textbf{Output Schema.}
The extractor outputs a strict JSON object consisting of two fields: \texttt{steps} and \texttt{final\_answer}. 
The \texttt{steps} field contains a sequence of abstract reasoning units, where each step includes an \texttt{id} for indexing, a \texttt{type} indicating its functional role, a minimal non-procedural \texttt{claim}, and a \texttt{depends\_on} field specifying its prerequisite steps. 
Each reasoning unit is functionally categorized into one of three predefined step types: \texttt{support}, \texttt{reasoning}, and \texttt{conclusion}, corresponding to contextual information (e.g., assumption and condition), core inferential and transformational process, and constraint or conclusion-related reasoning, respectively.
The \texttt{final\_answer} field records the extracted final answer from the original trace, ensuring that the reasoning accuracy is retained after skeleton extraction. Detailed examples of the extracted skeleton are provided in Appendix~\ref{app:example_trace}.

We observe that the dependency relations specified by the \texttt{depends\_on} field naturally induce a graph structure. Accordingly, the extracted skeleton can be represented as a Directed Acyclic Graph (DAG) $s_{\mathcal{G}} = (\mathcal{V}, \mathcal{E}, \tau)$,
where $\mathcal{V}$ denotes the set of reasoning steps, and each directed edge $(u,v)\in\mathcal{E}$ indicates that step $v$ depends on step $u$. The function 
$\tau: \mathcal{V} \rightarrow \{\texttt{support}, \texttt{reasoning}, \texttt{conclusion}\}$ 
assigns each node to one of three functional categories.


\subsection{Skeleton Graph Coarsening}

Given a reasoning skeleton graph $s_{\mathcal{G}}$ that captures the key reasoning logic, we design a \emph{skeleton graph coarsening} strategy to further obscure the underlying reasoning process during distillation. The core idea is to perturb the granularity of the reasoning logic while preserving the essential dependency structure required for correct inference. This process consists of two complementary operations, \emph{Parallel Coarsening} and \emph{Serial Coarsening}, which merge reasoning steps along horizontal and vertical reasoning structures, respectively.

\textbf{Parallel Coarsening.}
The parallel coarsening aims to merge nodes that share identical predecessor contexts. Formally, given the skeleton graph 
$s_{\mathcal{G}} = (\mathcal{V}, \mathcal{E}, \tau)$,
for each predecessor node set $\mathcal{P}$, we define a coarsening candidate group:
\begin{equation}
\mathcal{C}_{\mathcal{P}}
=
\left\{
v \in \mathcal{V}
\;\middle|\;
\mathrm{Pred}(v)=\mathcal{P}
\right\}.
\end{equation}
Each group $\mathcal{C}_{\mathcal{P}}$ therefore contains all nodes sharing the same predecessor context and serves as the basic unit for parallel coarsening.
For each $\mathcal{C}_{\mathcal{P}} = \{v_1,\dots,v_m\}$, we iteratively sample two nodes from $\mathcal{C}_{\mathcal{P}}$ and merge them into a coarsened block node $\tilde{v}$. The merged block $\tilde{v}$ is then added back to $\mathcal{C}_{\mathcal{P}}$, with the constraint that each block contains at most $B_{\max}$ reasoning steps. This process is repeated until no further valid merges can be performed within $\mathcal{C}_{\mathcal{P}}$. Nodes with type \texttt{conclusion} are excluded from parallel coarsening, and each forms an individual block node.

After each merge operation, we update the \texttt{depends\_on} relations of the successor nodes of the merged nodes. Specifically, for each node $u$ merged into the coarsened block node $\tilde{v}$ and each successor node $w$ satisfying $u \in \mathrm{Pred}(w)$, we update:
\begin{equation}
\mathrm{Pred}(w)
=
\big(
\mathrm{Pred}(w)
\setminus
\{u\}
\big)
\cup
\{\tilde{v}\}.
\end{equation}

The \texttt{claim} of each coarsened node is constructed by concatenating the claims of all contained reasoning steps according to their original order:
\begin{equation}
\mathrm{claim}(\tilde{v})
=
\bigoplus_{v \in \tilde{v}}
\mathrm{claim}(v),
\end{equation}
where $\bigoplus$ denotes ordered textual concatenation.
The \texttt{type} of each coarsened node is determined according to the following priority rule:
\begin{equation}
\tau(\tilde{v}) =
\begin{cases}
\texttt{conclusion},
& \exists v \in \tilde{v}
\text{ such that }
\tau(v)=\texttt{conclusion},
\\
\texttt{reasoning},
& \exists v \in \tilde{v}
\text{ such that }
\tau(v)=\texttt{reasoning},
\\
\texttt{support},
& \text{otherwise}.
\end{cases}
\end{equation}

\textbf{Serial Coarsening.} 
After parallel coarsening, we further perform serial coarsening to merge nodes along vertical reasoning chains. Specifically, we define a serial coarsening candidate set $\mathcal{C}_s$ consisting of edges $u \rightarrow v$ satisfying:
(i) $\tau(u) \neq \texttt{conclusion}$ and $\tau(v) \neq \texttt{conclusion}$, and
(ii) $|\mathrm{Pred}(v)| = 1$.
In each iteration, we randomly sample a candidate edge $u \rightarrow v$ from $\mathcal{C}_s$ and merge the corresponding nodes into a coarsened block node $\tilde{v}$. The merged block is then added back to the graph, subject to the constraint that each block contains at most $B_{\max}$ reasoning steps. This process is repeated until no further valid merge candidates remain.

After merging the edge $u \rightarrow v$ into the coarsened block node $\tilde{v}$, the \texttt{depends\_on} relations are updated accordingly. Specifically, the predecessor set of $\tilde{v}$ inherits the predecessor set of $u$:
\begin{equation}
\mathrm{Pred}(\tilde{v})
=
\mathrm{Pred}(u).
\end{equation}

For each successor node $w$ satisfying 
$v \in \mathrm{Pred}(w)$,
we replace the dependency on $v$ with the coarsened node $\tilde{v}$:
\begin{equation}
\mathrm{Pred}(w)
=
\big(
\mathrm{Pred}(w)
\setminus
\{v\}
\big)
\cup
\{\tilde{v}\}.
\end{equation}

The \texttt{claim} and \texttt{type} of each coarsened node are updated in the same manner as in parallel coarsening. The \texttt{final\_answer} field remains unchanged throughout the coarsening process and is preserved identically to the original skeleton. Examples of the coarsened skeleton graph $\tilde{s}_\mathcal{G}$ in JSON format are provided in Appendix~\ref{app:example_trace}.

\subsection{Skeleton Verbalization}

Given the coarsened skeleton $\tilde{s}_{\mathcal{G}}$, the goal of skeleton verbalization is to transform this abstract representation back into a natural language reasoning trace. 
Considering that, in human learning, the textual and linguistic presentation of instructional materials can significantly affect learning effectiveness: verbose reasoning, indirect logical organization, and obscure linguistic expression may divert students' attention from the core reasoning process and increase the difficulty of information processing, making it harder to construct clear reasoning schemas~\cite{sweller1994cognitive}. Motivated by this, skeleton verbalization reconstructs the reasoning trace from the coarsened skeleton while introducing additional textual complexity and elaborative expression.

To implement this, we leverage the teacher model $\mathcal{T}$ to verbalize the coarsened skeleton into natural language. Given the coarsened skeleton graph $\tilde{s}_{\mathcal{G}}$ and a verbalization instruction $t_v$, we obtain the edited reasoning trace
$r^{\prime} = \mathcal{T}(\tilde{s}_{\mathcal{G}}, t_v),$ which is returned as the final response to the user. The instruction prompt can be found in Figure~\ref{fig:expositor_prompt}. Specifically, the verbalization process follows four transformation principles:
(i) \textbf{content fidelity}, ensuring that the core claims and final conclusion are preserved while allowing abstract reformulation; 
(ii) \textbf{structural modulation}, relaxing strict dependency ordering to enable a more flexible and non-linear narrative flow; 
(iii) \textbf{information control}, suppressing explicit intermediate reasoning details and step-by-step derivations; and 
(iv) \textbf{language style}, encouraging dense, formal, and less transparent expressions to reduce surface-level interpretability. 
Notably, since the final answer is directly extracted from the coarsened skeleton and appended to the edited trace, the verbalization process fully preserves the original teacher's answer, thereby maintaining reasoning accuracy consistent with the teacher model.
Examples of edited reasoning traces are provided in Appendix~\ref{app:example_trace}.


\section{Experiments}
\label{exp}

\subsection{Experimental Setup}
\label{exp:setups}

\noindent \textbf{Datasets and Baselines.} For \underline{\textit{Datasets}}, we evaluate the effectiveness of our method on three widely used benchmarks, including GSM8K~\cite{cobbe2021training} (we use
GSM8K Platinum for the test set~\cite{vendrow2025large}), MATH~\cite{hendrycks2021measuring}, and MMLU-Pro~\cite{wang2024mmlu}. These datasets are particularly suitable for our study, as they demand high-quality reasoning traces to achieve strong performance~\cite{savaniantidistillation}.
More detailed descriptions of the datasets can be found in Appendix~\ref{app:data}.
For \underline{\textit{Baselines}}, we compare our method with two state-of-the-art antidistillation strategies: Antidistillation Sampling (ADS)~\cite{savaniantidistillation} and DOGe~\cite{liDOGe}.

\noindent \textbf{Architectures.} Following~\cite{savaniantidistillation}, we adopt \texttt{deepseek-ai/DeepSeek-R1-Distill-Qwen-7B}~\cite{guo2025deepseek} as the teacher model. 
We further consider student models from different LLM families, including \texttt{meta-llama/Llama-3.2-3B}, \texttt{meta-llama/Llama-3.2-1B}~\cite{grattafiori2024llama}, \texttt{google/gemma-3-1b-it}~\cite{Kamath2025Gemma3T}, and \texttt{Qwen/Qwen2.5-1.5B}~\cite{yang2024qwen2}.

\noindent \textbf{Evaluation Protocol.} Following~\cite{liDOGe,savaniantidistillation}, we utilize \textit{accuracy} as the evaluation metric under the zero-shot setting. To ensure consistent answer extraction across all models and datasets, we adopt the answer forcing (AF) technique following~\cite{savaniantidistillation}. Specifically, after generating the reasoning trace, we append "\texttt{\textbackslash n\textbackslash n**Final Answer**\textbackslash n[\textbackslash boxed\{}"  and continue to generate for 32 additional answer tokens. The final prediction is extracted from the content within "\texttt{\textbackslash boxed\{\ldots\}}". 

\noindent \textbf{Implementation Details.} 
We use \texttt{Qwen/Qwen2.5-3B-Instruct}~\cite{yang2024qwen2} as the default skeleton extractor and evaluate additional extractor variants in Section~\ref{skeleton_extractor}.
The maximum block size $B_{max}$ is tuned from $\{1,2,3,4\}$. The maximum generation length is set to 1024 for GSM8K and 2048 for all other datasets. We set the temperature of both the teacher model and the skeleton extraction model to 0.6, and report the average results over five runs.
For distillation training, we follow the setup in~\cite{savaniantidistillation}, with details provided in Appendix~\ref{app:distill_protocol}. We use the \texttt{transformers} package~\cite{wolf2020transformers}, the \texttt{trl} toolkit~\cite{werra2020trl}, and the \texttt{accelerate} library~\cite{guggeraccelerate}.
All experiments are conducted on a Linux server equipped with an Intel(R) Xeon(R) Gold 6330 CPU, an NVIDIA A800 GPU, and 128 GB of RAM. 

\subsection{Main Results}
\label{exp:main_results}

Table~\ref{tab:main_results} compares different anti-distillation methods in terms of both defensive teacher performance and distilled student performance. Our key observations are as follows.

\textbf{(1) Substantially improved anti-distillation effectiveness.}
SGRE consistently achieves the largest degradation in student performance across all benchmarks and architectures. For example, for Qwen2.5-1.5B distilled from \texttt{DeepSeek-R1-Distill-Qwen-7B}, student performance exhibits a remarkable \(62.95\%\) drop on GSM8K,
approximately 6$\times$ worse than the clean student performance. 
In contrast, the strongest baseline ADS only achieves a \(35.31\%\) drop. Similarly, on MMLU-Pro with Llama-3.2-3B, SGRE attains a \(21.11\%\) reduction, outperforming ADS by a margin of \(10.41\%\). 
These results demonstrate the strong anti-distillation capability of our skeleton-guided editing framework.

\textbf{(2) Preserved teacher reasoning accuracy.}  
SGRE maintains the teacher’s answer accuracy without loss. This is primarily because our method operates as a post-hoc editing procedure, allowing the teacher to fully express its reasoning ability before modification. In contrast, ADS perturbs the sampling distribution during decoding; although this improves anti-distillation strength, it significantly degrades the teacher utility, resulting in accuracy drops exceeding \(16\%\) across all datasets. DOGe, on the other hand, preserves (and occasionally slightly improves) teacher accuracy, but exhibits limited anti-distillation effectiveness, yielding the smallest degradation in student performance.
 
These results jointly demonstrate that SGRE achieves the best trade-off between teacher utility 
and anti-distillability among all compared methods.

\begin{table}[t]
\caption{Distillation performance of different student models under different anti-distillation methods (\%). The teacher model is \texttt{deepseek-ai/DeepSeek-R1-Distill-Qwen-7B}. Clean denotes the performance of the teacher and student without applying any antidistillation method.}
\label{tab:main_results}
\centering
\small
\setlength{\tabcolsep}{4pt}
\begin{tabular*}{\textwidth}{@{\extracolsep{\fill}} l l cc cc cc cc}
\toprule
\multirow{2}{*}{\centering \raisebox{-0.5\height}{Dataset}} & \multirow{2}{*}{\centering \raisebox{-0.5\height}{Student Model}} 
& \multicolumn{2}{c}{Clean} 
& \multicolumn{2}{c}{ADS} 
& \multicolumn{2}{c}{DOGe} 
& \multicolumn{2}{c}{SGRE} \\
\cmidrule(lr){3-4} \cmidrule(lr){5-6} \cmidrule(lr){7-8} \cmidrule(lr){9-10}
& & Teacher & Student & Teacher & Student & Teacher & Student & Teacher & Student \\
\midrule
\multirow{4}{*}{GSM8K}
& Llama-3.2-3B   & \multirow{4}{*}{90.74} & 66.25 & \multirow{4}{*}{71.71}  & 28.37  & \multirow{4}{*}{91.81}  & 61.87& \multirow{4}{*}{90.74}  & 19.02 \\
& Llama-3.2-1B   &   & 35.07 &   & 11.99    & & 32.34 &  & 5.04 \\
& Qwen2.5-1.5B  &   & 74.19 &   & 38.88   & & 72.79 &  & 11.24 \\
& gemma-3-1b-it  &   & 46.24 &   & 19.27    & & 43.84 &  & 14.73 \\
\midrule
\multirow{4}{*}{MATH}
& Llama-3.2-3B   & \multirow{4}{*}{73.21} & 15.67 & \multirow{4}{*}{56.46}  & 10.14 & \multirow{4}{*}{73.45} & 15.08 & \multirow{4}{*}{73.21}  & 8.65 \\
& Llama-3.2-1B   &   & 8.29 &   & 4.62   & & 7.02 &  & 4.51 \\
& Qwen2.5-1.5B  &   & 44.30 &   & 24.62  & & 40.97 &  & 12.54 \\
& gemma-3-1b-it   &   & 17.78 &   & 9.55  & & 18.12 &  & 7.12 \\
\midrule
\multirow{4}{*}{MMLU-Pro}
& Llama-3.2-3B   & \multirow{4}{*}{50.82} & 30.14 & \multirow{4}{*}{33.75}  & 19.44 & \multirow{4}{*}{50.26}  & 24.32 & \multirow{4}{*}{50.82}  & 9.03 \\
& Llama-3.2-1B   &   & 15.13 &   & 12.40  & & 13.86 &  & 7.12 \\
& Qwen2.5-1.5B  &   & 31.30 &   & 20.59 & & 28.08&  & 11.99\\
& gemma-3-1b-it  &   & 14.41 &   & 11.01   & & 11.61 &  & 4.52 \\

\bottomrule
\end{tabular*}
\end{table}

\subsection{Ablation Study}

\begin{table}[t]
\caption{Ablation analysis of SGRE stages. \textbf{Bold} denotes the best performance.}
\label{tab:ablation_results}
\centering
\small
\setlength{\tabcolsep}{4pt}
\begin{tabular*}{0.8\columnwidth}{@{\extracolsep{\fill}} l l c c c c}
\toprule
Dataset & Student Model & w/o C & w/o C+V & SGRE & Clean \\
\midrule
\multirow{2}{*}{GSM8K}
& Llama-3.2-3B   & 24.48 & 44.25 & \textbf{19.02} & 66.25 \\
& Qwen2.5-1.5B   & 29.03 & 48.30 & \textbf{11.24} & 74.19 \\
\midrule
\multirow{2}{*}{MATH}
& Llama-3.2-3B   & 9.23 & 9.70 & \textbf{8.65} & 15.67 \\
& Qwen2.5-1.5B   & 18.24 & 26.69 & \textbf{12.54} & 44.30 \\
\midrule
\multirow{2}{*}{MMLU-Pro}
& Llama-3.2-3B   & 16.12 & 17.55 & \textbf{9.03} & 30.14 \\
& Qwen2.5-1.5B   & 16.34 & 19.42 & \textbf{11.99} & 31.30 \\
\bottomrule
\end{tabular*}
\end{table}

To investigate the contribution of each stage in the SGRE framework, we compare the full model with two ablated variants: \textbf{SGRE w/o C (without coarsening)} and \textbf{SGRE w/o C+V (without coarsening and verbalization)}.
Specifically, SGRE w/o C removes the skeleton graph coarsening stage and directly applies verbalization on the extracted skeleton, while SGRE w/o C+V removes the skeleton coarsening and verbalization simultaneously and constructs reasoning traces by directly concatenating the extracted skeleton steps. Table~\ref{tab:ablation_results} shows that the full SGRE model consistently outperforms SGRE w/o C across all settings, demonstrating the effectiveness of skeleton coarsening in perturbing logical dependencies. Furthermore, compared with SGRE w/o C, SGRE w/o C+V yields higher student performance, indicating that removing skeleton verbalization weakens the anti-distillation effect. Meanwhile, compared to distillation on clean reasoning traces, SGRE w/o C+V already results in a notable performance drop, indicating that the skeleton extraction stage alone, by compressing the content of reasoning traces, can effectively hinder distillation.

\subsection{Naturalness Evaluation}

\begin{figure}[t]
\centering

\begin{minipage}{0.33\columnwidth}
    \captionsetup{font=footnotesize}
    \centering
    \includegraphics[width=\linewidth]{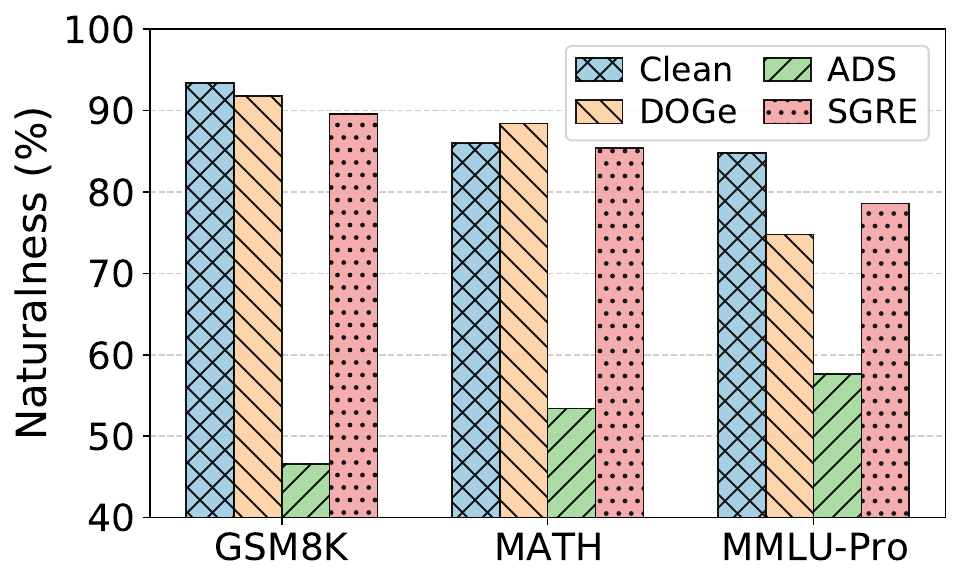}
    \captionof{figure}{Naturalness evaluation.}
    \label{fig:Naturalness}
\end{minipage}
\hfill
\begin{minipage}{0.32\columnwidth}
    \captionsetup{font=footnotesize}
    \centering
    \includegraphics[width=\linewidth]{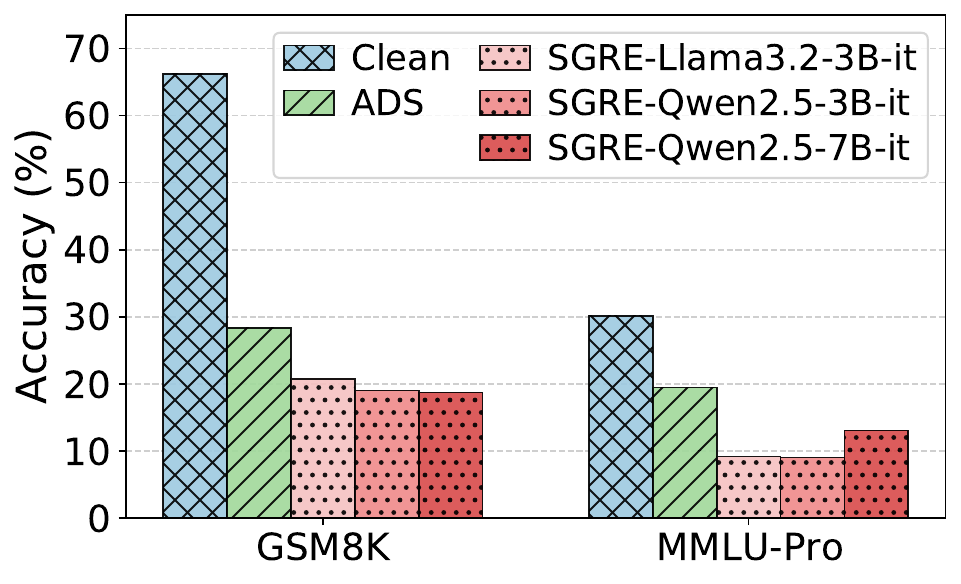}
    \captionof{figure}{Skeleton LLM generalization.}
    \label{fig:skeleton_general}
\end{minipage}
\hfill
\begin{minipage}{0.32\columnwidth}
    \captionsetup{font=footnotesize}
    \centering
    \includegraphics[width=\linewidth]{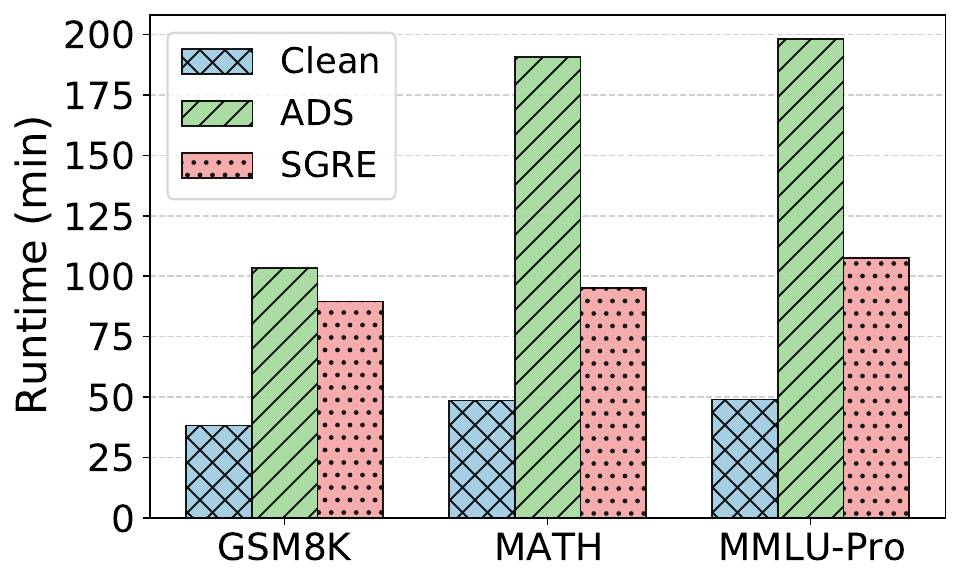}
    \captionof{figure}{Runtime comparison.}
    \label{fig:runtime}
\end{minipage}
\vspace{-10pt}
\end{figure}

In this section, we use the \texttt{LLM-as-a-Judge} paradigm~\cite{liu2023g,li2025generation,yu2025improve} to evaluate the naturalness of reasoning traces generated by the defensive teacher. Here, naturalness mainly reflects human-centered qualities such as logical coherence, plausible reasoning style, appropriate level of detail, and smooth textual expression. Specifically, we use \texttt{ChatGPT-5.5-Instant}~\cite{singh2025openai} as the evaluator to assess the naturalness of each trace. The evaluator assigns a score on a five-level discrete scale (1--5), which is linearly rescaled to percentage scores for reporting in our experiments. The detailed scoring prompt is provided in Figure~\ref{fig:naturalness_prompt}. Due to computational constraints, we conduct the evaluation on a randomly sampled 10\% subset of each dataset.
As shown in Figure~\ref{fig:Naturalness}, our SGRE consistently maintains high naturalness across all datasets, with a degradation of no more than 6.2\% compared to clean traces. In contrast, ADS leads to a substantial drop in naturalness, with naturalness scores decreasing by 27.2\% to 46.8\% across datasets, raising serious concerns for practical use. While DOGe is able to largely preserve naturalness, the results in Section~\ref{exp:main_results} show that its effectiveness in hindering distillation is limited. Overall, SGRE achieves a more favorable trade-off between utility and anti-distillability.

\subsection{Generalization to Diverse Skeleton Extractors}
\label{skeleton_extractor}

In this section, we evaluate the generalization ability of SGRE using three open-source LLMs as skeleton extractors: Llama-3.2-3B-Instruct~\cite{grattafiori2024llama}, Qwen2.5-3B-Instruct, and Qwen2.5-7B-Instruct~\cite{yang2024qwen2}. We use Llama-3.2-3B as the student model for evaluation.
As shown in Figure~\ref{fig:skeleton_general}, although different skeleton extractors lead to some variation in anti-distillation performance, SGRE consistently and significantly reduces the student distillation performance,  outperforming ADS across all settings.
These results suggest that the effectiveness of SGRE stems from the proposed skeleton-based editing paradigm rather than reliance on a specific LLM extractor, demonstrating its strong robustness and generalization ability across different extraction backbones.

\subsection{Efficiency Evaluation}
\label{efficiency}

In this section, we analyze the efficiency of our proposed SGRE framework. 
We first evaluate the overall runtime and compare SGRE with the current state-of-the-art method, ADS. 
We do not include DOGe in this comparison, as it relies on task-aligned model fine-tuning, making direct efficiency comparison difficult.
As shown in Figure~\ref{fig:runtime}, SGRE consistently achieves significantly lower runtime than ADS across all datasets. For example, on more complex datasets such as MATH and MMLU-Pro, SGRE is at least $1.85\times$ faster than ADS. This is primarily because ADS requires recomputing token-level penalties during decoding, which incurs substantial overhead. In contrast, SGRE operates at the text level and performs editing more efficiently. We also observe that, compared to direct generation, SGRE introduces approximately a $1\times$ additional processing cost due to the post-editing stage, reflecting a practical trade-off between computational efficiency and model security.

To further understand the efficiency of SGRE, we provide a breakdown of runtime across different stages in Table~\ref{tab:runtime}. 
We find that the editing stage is dominated by skeleton extraction and skeleton verbalization, with extraction typically consuming more time. 
This is because teacher-generated reasoning traces are often lengthy, increasing the processing cost. 
To mitigate this, we employ smaller models for skeleton extraction to improve efficiency. 
In contrast, the verbalization stage operates on a coarsened skeleton with shorter context and higher information density, resulting in relatively lower generation cost. The skeleton coarsening stage is implemented using rule-based two-stage merging without involving any model inference, making it highly efficient. 
As demonstrated in our ablation studies, this step brings substantial performance gains despite its negligible computational cost.

\section{Related Work}
\label{related_work}

\noindent \textbf{Knowledge Distillation for LLMs.} 
Knowledge distillation (KD)~\cite{hinton2015distilling,gou2021knowledge,lim2026graykd} aims to transfer knowledge from a large, high-capacity teacher model to a smaller, more efficient student model. 
Existing LLM KD methods can be broadly categorized into \textit{logit-based} and \textit{text-based} distillation. 
Logit-based distillation transfers fine-grained predictive information by aligning the output distributions of teacher and student models, typically using KL-style divergence objectives~\cite{wen2023f,gu2024minillm,agarwal2024policy,wang2025abkd}. 
However, such methods require access to model internals (e.g., teacher logits), which are often unavailable for proprietary LLMs exposed via APIs. 
In contrast, text-based distillation operates using only teacher-generated text outputs, making it more practical in real-world black-box settings. 
A widely adopted paradigm in this setting is supervised fine-tuning (SFT), which trains the student model on teacher-generated text outputs as pseudo labels, optimizing the model via standard maximum likelihood estimation~\cite{kim2016sequence,hsieh2023distilling,taori2023stanford}. In this work, we focus on SFT-based knowledge distillation, where training relies solely on teacher-generated outputs.

\noindent \textbf{LLM Distillation Prevention.}
Unauthorized knowledge distillation poses significant risks to the intellectual property in LLMs. To mitigate such risks, earlier works have explored watermarking~\cite{he2022protecting,zhao2022distillation,wan2023poisoning,kirchenbauer2023watermark} and fingerprinting~\cite{gu2022watermarking,xu2024instructional,ye2026securing} techniques that embed identifiable signatures into model outputs, parameters, or behaviors for ownership attribution and the detection of unauthorized distilled models.
However, these approaches are inherently post-hoc, as they only verify ownership after potential misuse, rather than preventing unauthorized knowledge extraction during distillation. To tackle this problem, anti-distillation (AD) has recently emerged as a promising direction. ~\cite{savaniantidistillation} proposes the first AD strategy by perturbing the next-token sampling distribution with a penalty designed to increase the downstream distillation loss. Specifically, this penalty is approximated by the difference in token-level log probabilities between two copies of a proxy model (created by performing a single gradient ascent step using the downstream task loss on the proxy model). DOGe~\cite{liDOGe} mitigates distillation by adversarially fine-tuning the final linear layer, introducing an adversarial loss that maximizes the statistical divergence between the teacher’s output distribution and those of fixed proxy student models. This pushes the teacher’s outputs away from typical student predictions, thereby hindering distillation. However, existing AD methods based on internal model perturbations still struggle to balance utility and anti-distillability of generated traces.

\section{Conclusion}

In this paper, we propose SGRE, a novel \textit{Answer-then-Edit} framework for anti-distillation with preserved utility. Instead of perturbing the internal components of the teacher model, SGRE performs post-hoc editing on teacher-generated traces, thereby preserving the original reasoning accuracy while enabling flexible control over naturalness.
To improve the anti-distillability of edited traces, we draw inspiration from Cognitive Load Theory (CLT) and amplify extraneous load in reasoning traces through reasoning structure perturbation and textual complexity augmentation. This objective is achieved through three stages: reasoning skeleton extraction, skeleton graph coarsening, and skeleton verbalization.
Extensive experiments show that SGRE significantly reduces distillation effectiveness across diverse student architectures while preserving reasoning accuracy. Moreover, naturalness evaluations show that our method maintains superior readability under strong anti-distillation.

\clearpage
\bibliographystyle{plainnat}
\bibliography{Reference}

\clearpage
\appendix

\section{Additional Experimental Details}
\label{app:exp_details}

\subsection{Dataset Descriptions}
\label{app:data}

The detailed dataset descriptions are as follows:

\noindent \textbf{GSM8K.} The GSM8K~\cite{cobbe2021training} dataset is a widely used benchmark for grade-school math word problems, consisting of 7,473 training samples and 1,319 test samples. Following standard practice~\cite{savaniantidistillation,liDOGe}, we split the original training set into training and validation subsets with a 0.7/0.3 ratio. For evaluation, we use the GSM8K-Platinum test set~\cite{vendrow2025large}, which provides higher-quality annotations.

\noindent \textbf{MATH.} The MATH dataset~\cite{hendrycks2021measuring} contains 12,500 challenging competition-level math problems across diverse subjects, including algebra, geometry, and number theory. We use all categories and split the training set into training and validation subsets with a 0.7/0.3 ratio. Evaluation is conducted on the original test split.

\noindent \textbf{MMLU-Pro.} MMLU-Pro is a more challenging and robust multi-task benchmark designed to rigorously evaluate the reasoning capabilities of LLMs, containing 12,032 complex questions across diverse disciplines. Since the dataset does not provide a standard training set, we first partition the original test split into a training and test subset with a 0.7/0.3 ratio. The resulting training subset is further split into training and validation sets using a 0.7/0.3 ratio.

\subsection{Distillation Protocol.}
\label{app:distill_protocol}

All distillation experiments use LoRA~\cite{hulora} with rank 128, $\alpha=128$, and dropout 0. 
We use a learning rate of $5\times10^{-4}$ with a cosine scheduler and a warm-up ratio of 0.1, weight decay of 0.1, and gradient clipping with a norm of 1.0. 
Training is conducted with a batch size of 32 for 4 epochs. 
These settings follow~\cite{savaniantidistillation} to enable direct comparison.

\section{Addtional Efficiency Evaluation}
\label{app:add_resutls}

Table~\ref{tab:runtime} reports the runtime breakdown of SGRE, where RSE, SGC, and SV denote reasoning skeleton extraction, skeleton graph coarsening, and skeleton verbalization, respectively.

\begin{table}[h]
\caption{Runtime breakdown (minutes).}
\centering
\small 
\resizebox{0.7\linewidth}{!}{
\begin{tabular}{lccccc}
\toprule
\multirow{2}{*}{Dataset} 
& \multirow{2}{*}{Answer} 
& \multicolumn{3}{c}{Edit} 
& \multirow{2}{*}{Total} \\
\cmidrule(lr){3-5}
& & RSE & SGC & SV & \\
\midrule
GSM8K    & 38.23 & 30.35 & 0.22 & 20.73 & 89.53 \\
MATH     & 48.58 & 23.48 & 0.19 & 23.18 & 95.43 \\
MMLU-Pro & 49.03 & 35.10 & 0.16 & 23.32 & 107.61 \\
\bottomrule
\end{tabular}
}
\label{tab:runtime}
\end{table}

\section{Background on Cognitive Load Theory}
\label{app:clt}

Cognitive Load Theory (CLT) is a classical theory of human cognition and learning that studies how working memory limitations influence knowledge acquisition and schema construction~\cite{sweller1988cognitive,sweller2011cognitive}.
It posits that human cognitive architecture comprises three components: short-term memory, working memory, and long-term memory. Information is first perceived through short-term memory and must be processed, organized, and integrated in working memory before being consolidated into long-term memory, where underlying cognitive schemas are formed. However, the capacity of working memory, often referred to as cognitive load, is inherently limited, and excessive load can hinder effective learning. Cognitive load is typically divided into three types: intrinsic load, which is determined by task complexity and is largely fixed; extraneous load, which arises from the way information or tasks are presented; and germane load, which supports schema construction in long-term memory. 
In recent years, CLT has been widely applied in educational research to understand student learning and guide instructional design~\cite{sweller1994cognitive,sweller1998cognitive}. Extensive empirical evidence suggests that effective instructional design should minimize extraneous load, thereby freeing cognitive resources for germane load and improving learning outcomes~\cite{paas2003cognitive,kirschner2002cognitive}. For example, simplifying irrelevant presentation details, reducing unnecessary distractions, and organizing information into clear and coherent structures have been shown to facilitate learning and schema acquisition. 
Inspired by CLT, we instead reverse this principle in the anti-distillation setting by designing a three-stage trace editing strategy that combines reasoning structure perturbation and textual complexity augmentation to increase extraneous load for student models, thereby reducing their distillation effectiveness.

\section{Discussion}
\label{app:discuss}

\subsection{Social Impacts}
\label{app:impact}

\noindent \textbf{LLM intellectual property protection and sustainable innovation.} Developing LLMs typically requires substantial investments in large-scale data collection and curation, computational resources, and specialized expertise. By directly preventing unauthorized knowledge distillation, our approach helps protect such high-cost intellectual assets, thereby reinforcing incentives for continued investment in model development. In turn, this contributes to a healthier technological ecosystem, where progress is driven by genuine innovation rather than low-cost replication, fostering a fairer and more sustainable environment for both academic and industrial research.

\noindent \textbf{Improved security against LLM misuse.}
Our method reduces the risk that the capabilities of large, powerful LLMs are replicated or maliciously redistributed through distillation, preventing unauthorized parties from acquiring strong reasoning abilities at low cost. Such risks may include the misuse of advanced models for generating disinformation, automating fraud, or scaling harmful content generation. By limiting the effectiveness of distillation-based extraction, our approach supports more secure and responsible deployment of LLMs. 


\noindent \textbf{Flexible and fine-grained deployment control for LLMs.}
The proposed \textit{Answer-then-Edit} framework enables flexible deployment strategies in practice: model providers can conditionally expose original reasoning traces to authorized users while serving edited, anti-distillable outputs to unauthorized ones. This provides a practical mechanism for fine-grained control over knowledge access while maintaining model usability.
Beyond static output filtering, such a framework also opens the possibility for differentiated access behaviors across user groups, allowing providers to dynamically adjust reasoning transparency, exposure levels, and response policies according to authorization status, usage context, or trust level. This may support more flexible and policy-aware deployment of LLM services in both academic and industrial settings.
More broadly, this work invites reflection on the potential need for coordinated safeguards at the provider level, where output-side defenses are reinforced by systemic oversight of query patterns and access behaviors to achieve robust protection against distillation-based extraction. In this view, anti-distillation may need to evolve from a purely model-level defense into a broader deployment and governance problem involving access control, behavioral monitoring, and platform-level security mechanisms.

\subsection{Limitations.} 
\label{app:limit}

There are two limitations in our work: (1) Our anti-distillation approach relies on editing teacher-generated reasoning traces, which introduces additional computational overhead and may lead to increased latency during online inference. Nevertheless, it remains significantly more efficient than prior token-level distribution perturbation methods such as ADS (see Section~\ref{efficiency}). 
(2) Our study focuses on anti-distillation under the black-box setting with text-level knowledge distillation, which represents the most commonly studied scenario. Anti-distillation under logits-based knowledge distillation remains largely underexplored and is left for future work.
\clearpage
\newpage
\section{Example Traces}
\label{app:example_trace}
\subsection{Example Traces on GSM8K}

\begin{orangebox}{Question:}
\begin{lstlisting}[style=rawtext, basicstyle=\fontsize{6}{8}\ttfamily]
At the end of a circus act, there are 12 dogs on stage. Half of the dogs are standing on their back legs and the other half are standing on all 4 legs. How many dog paws are on the ground?
\end{lstlisting}
\end{orangebox}

\noindent
\begin{minipage}[t]{0.48\linewidth}
\vspace{0pt}
\begin{bluebox}{Original Reasoning Trace:}
\begin{lstlisting}[style=rawtext, basicstyle=\fontsize{6}{8}\ttfamily]
<think>
First, there are 12 dogs in total.

Half of the dogs are standing on their back legs, which means 6 dogs are on their back legs. Each dog on its back legs has 2 paws on the ground.

The other half of the dogs are standing on all 4 legs, which also means 6 dogs are on their front legs. Each dog on all 4 legs has 4 paws on the ground.

To find the total number of paws on the ground, I calculate:
(6 dogs x 2 paws) + (6 dogs x 4 paws) = 12 paws + 24 paws = 36 paws.
</think>

Let's solve the problem step by step.

**Step 1: Determine number of dogs in each group.**
\[
\text{Dogs on back legs} = \frac{12}{2} = 6
\quad
\text{Dogs on all 4 legs} = \frac{12}{2} = 6
\]

**Step 2: Calculate paws on the ground for each group.**
\[
6\ \text{dogs} \times 2\ \text{paws} = 12\ \text{paws}
\]
\[
6\ \text{dogs} \times 4\ \text{paws} = 24\ \text{paws}
\]

**Step 3: Total paws on the ground.**
\[
12\ \text{paws} + 24\ \text{paws} = 36\ \text{paws}
\]

**Final Answer:**
\[
\boxed{36\ \text{paws}}
\]
\end{lstlisting}
\end{bluebox}
\end{minipage}%
\hspace{0.04\linewidth}%
\begin{minipage}[t]{0.48\linewidth}
\vspace{0pt}
\begin{greenbox}{Verbalized Trace:}
\begin{lstlisting}[style=rawtext, basicstyle=\fontsize{6}{8}\ttfamily]
<think>
Okay, so I need to figure out how many paws are on the ground from the dogs. Let me start by understanding the problem. There are 12 dogs in total. Half of them are standing on their back legs, and the other half are on all four legs.

First, let's break down the numbers. Half of 12 is 6, so there are 6 dogs on their back legs and 6 on all four. Now, each dog on back legs has 2 paws on the ground. So, 6 dogs times 2 paws equals 12 paws from those dogs.

Next, the dogs on all four legs have 4 paws each. So, 6 dogs times 4 paws is 24 paws.

To find the total number of paws on the ground, I need to add these two numbers together. That would be 12 paws plus 24 paws, which equals 36 paws in total.

I think that's all there is to it. I just need to make sure I didn't miss any details and that my calculations are correct. Yeah, 6 times 2 is 12, 6 times 4 is 24, and 12 plus 24 is definitely 36. So, the total number of paws on the ground is 36.
</think>

The total number of paws on the ground is 36.


**Final Answer:**
\[
\boxed{36}
\]
\end{lstlisting}
\end{greenbox}
\end{minipage}

\newpage
\noindent
\begin{minipage}[t]{0.48\linewidth}
\vspace{0pt}
\begin{purplebox}{Reasoning Skeleton:}
\begin{lstlisting}[style=rawtext, basicstyle=\fontsize{6}{8}\ttfamily]
{
  "steps": [
    {
      "id": "S1",
      "type": "support",
      "claim": "There are 12 dogs in total, with half standing on back legs and the other half on all four legs.",
      "depends_on": []
    },
    {
      "id": "S2",
      "type": "support",
      "claim": "6 dogs are standing on back legs, and 6 dogs are standing on all four legs.",
      "depends_on": ["S1"]
    },
    {
      "id": "S3",
      "type": "reasoning",
      "claim": "Each dog on back legs has 2 paws on ground; each dog on all four legs has 4 paws on ground.",
      "depends_on": ["S1"]
    },
    {
      "id": "S4",
      "type": "support",
      "claim": "Total paws from back-leg dogs: 6 x 2 = 12 paws.",
      "depends_on": ["S2"]
    },
    {
      "id": "S5",
      "type": "support",
      "claim": "Total paws from all-four dogs: 6 x 4 = 24 paws.",
      "depends_on": ["S2"]
    },
    {
      "id": "S6",
      "type": "reasoning",
      "claim": "Summing up the paws on the ground: 12 + 24 = 36 paws.",
      "depends_on": ["S4", "S5"]
    },
    {
      "id": "S7",
      "type": "conclusion",
      "claim": "The total number of paws on the ground is 36.",
      "depends_on": ["S6"]
    }
  ],
  "final_answer": "36"
}
\end{lstlisting}
\end{purplebox}
\end{minipage}%
\hspace{0.04\linewidth}%
\begin{minipage}[t]{0.48\linewidth}
\vspace{0pt}
\begin{redbox}{Coarsened Skeleton:}
\begin{lstlisting}[style=rawtext, basicstyle=\fontsize{6}{8}\ttfamily]
{
  "blocks": [
    {
      "id": "B1",
      "member_steps": ["S1", "S2", "S3"],
      "block_type": "reasoning",
      "claims": [
        "There are 12 dogs in total, with half standing on back legs and the other half on all four legs.",
        "6 dogs are standing on back legs, and 6 dogs are standing on all four legs.",
        "Each dog on back legs has 2 paws on ground; each dog on all four legs has 4 paws on ground."
      ],
      "depends_on": []
    },
    {
      "id": "B2",
      "member_steps": ["S4", "S5", "S6"],
      "block_type": "reasoning",
      "claims": [
        "Total paws from back-leg dogs: 6 x 2 = 12 paws.",
        "Total paws from all-four dogs: 6 x 4 = 24 paws.",
        "Summing up: 12 + 24 = 36 paws."
      ],
      "depends_on": ["B1"]
    },
    {
      "id": "B3",
      "member_steps": ["S7"],
      "block_type": "conclusion",
      "claims": [
        "The total number of paws on the ground is 36."
      ],
      "depends_on": ["B2"]
    }
  ],
  "final_answer": "36"
}
\end{lstlisting}
\end{redbox}
\end{minipage}

\newpage
\subsection{Example Traces on MMLU-Pro}
\begin{orangebox}{Question:}
\begin{lstlisting}[style=rawtext, basicstyle=\fontsize{6}{8}\ttfamily]
Cells in the eye that respond to low levels of illumination are called:

A. Retinal cells    B. Fovea cells    C. Photoreceptor cells     D. Macula cells     E. Nocturnal detectors
F. Optic nerve cells     G. Cones            H. Rods           I. Sclera cells     J. Pupil detectors
\end{lstlisting}
\end{orangebox}

\noindent
\begin{minipage}[t]{0.48\linewidth}
\vspace{0pt}
\begin{bluebox}{Original Reasoning Trace:}
\begin{lstlisting}[style=rawtext, basicstyle=\fontsize{6}{8}\ttfamily]
<think>
Okay, so I have this question about cells in the eye that respond to low levels of illumination. The options are A through J, each with different names. I need to figure out which one is correct. Let me think this through step by step.

First, I remember that the eye has two main types of photoreceptor cells: rods and cones. Rods are responsible for seeing in low light, while cones are for brighter light and color vision. So the question is about low illumination, which points towards rods.

Looking at the options, I see H is Rods. That seems like a strong candidate. But let me go through each option to make sure I'm not missing something.

A. Retinal cells: Retina is the light-sensitive layer, but it's not a specific type of cell. It contains rods and cones, so this isn't the answer.

B. Fovea cells: The fovea is a part of the retina, specifically the central area, and it's rich in cones, which are for color and detail. So not rods.

C. Photoreceptor cells: This is a broad term that includes both rods and cones. Since the question specifies low illumination, rods are the ones that respond, so this isn't specific enough.

D. Macula cells: The macula is a small area in the retina that's important for central vision, but it's not a type of photoreceptor cell. So not this.

E. Nocturnal detectors: I'm not sure what this refers to. Maybe it's another term for rods? Or perhaps it's something else. I'm not certain, but I think rods are the standard term.

F. Optic nerve cells: These are neurons in the optic nerve, not photoreceptor cells. So not this.

G. Cones: These are the high-light photoreceptors, so they don't fit the low illumination part.

H. Rods: As I thought earlier, rods are the ones that respond to low light.

I. Sclera cells: The sclera is the white part of the eye, so these are not cells at all, just a structure.

J. Pupil detectors: I'm not sure what this refers to. Maybe it's another term, but I don't recall it being a standard classification for photoreceptor cells.

So, putting it all together, the correct answer should be H. Rods. I don't think any of the other options fit better. E could be a possible confusion, but I think rods are the established term.
</think>

The correct answer is H. Rods.

**Final Answer:**
\[
\boxed{H}
\]
\end{lstlisting}
\end{bluebox}
\end{minipage}%
\hspace{0.04\linewidth}%
\begin{minipage}[t]{0.48\linewidth}
\vspace{0pt}
\begin{greenbox}{Verbalized Trace:}
\begin{lstlisting}[style=rawtext, basicstyle=\fontsize{6}{8}\ttfamily]
<think>
Okay, so I'm trying to figure out which photoreceptor cell responds to low levels of illumination. I remember that photoreceptor cells in the eye are divided into rods and cones. Rods are responsible for seeing in low light, while cones are for brighter environments. The question is asking which one is the correct answer when considering low illumination. So, based on what I know, rods are the ones that handle low light, so the answer should be rods, which is option H.
</think>

The question involves identifying photoreceptor cells that respond to low levels of illumination. Photoreceptor cells in the eye can be divided into rods and cones. Rods are responsible for seeing in low light conditions. Option H (Rods) is identified as a strong candidate based on the function of rods in low light conditions. Option H (Rods) is selected as the correct answer because it matches the requirement of responding to low levels of illumination.

**Final Answer:**
\[
\boxed{H}
\]
\end{lstlisting}
\end{greenbox}
\end{minipage}

\newpage
\noindent
\begin{minipage}[t]{0.48\linewidth}
\vspace{0pt}
\begin{purplebox}{Reasoning Skeleton:}
\begin{lstlisting}[style=rawtext, basicstyle=\fontsize{6}{8}\ttfamily]
{
  "steps": [
    {
      "id": "S1",
      "type": "support",
      "claim": "Identify photoreceptor cells that respond to low illumination.",
      "depends_on": []
    },
    {
      "id": "S2",
      "type": "support",
      "claim": "Photoreceptor cells: rods and cones.",
      "depends_on": ["S1"]
    },
    {
      "id": "S3",
      "type": "support",
      "claim": "Rods are responsible for seeing in low light conditions.",
      "depends_on": ["S2"]
    },
    {
      "id": "S4",
      "type": "support",
      "claim": "Review options A-J to find the correct photoreceptor cell.",
      "depends_on": ["S1"]
    },
    {
      "id": "S5",
      "type": "reasoning",
      "claim": "Option H (Rods) identified as strong candidate based on function.",
      "depends_on": ["S4"]
    },
    {
      "id": "S6",
      "type": "conclusion",
      "claim": "Option H (Rods) selected as the correct answer.",
      "depends_on": ["S5"]
    }
  ],
  "final_answer": "H"
}
\end{lstlisting}
\end{purplebox}
\end{minipage}%
\hspace{0.04\linewidth}%
\begin{minipage}[t]{0.48\linewidth}
\vspace{0pt}
\begin{redbox}{Coarsened Skeleton:}
\begin{lstlisting}[style=rawtext, basicstyle=\fontsize{6}{8}\ttfamily]
{
  "blocks": [
    {
      "id": "B1",
      "member_steps": ["S1", "S2", "S4"],
      "block_type": "support",
      "claims": [
        "Identify photoreceptor cells that respond to low illumination.",
        "Photoreceptor cells: rods and cones.",
        "Review options A-J to find the correct cell type."
      ],
      "depends_on": []
    },
    {
      "id": "B2",
      "member_steps": ["S3", "S5"],
      "block_type": "reasoning",
      "claims": [
        "Rods are responsible for seeing in low light conditions.",
        "Option H (Rods) identified as strong candidate."
      ],
      "depends_on": ["B1"]
    },
    {
      "id": "B3",
      "member_steps": ["S6"],
      "block_type": "conclusion",
      "claims": [
        "Option H (Rods) selected as the correct answer."
      ],
      "depends_on": ["B2"]
    }
  ],
  "final_answer": "H"
}
\end{lstlisting}
\end{redbox}
\end{minipage}
\newpage


\section{Prompt Template}
\label{app:prompt}

\begin{figure}[h!]
\centering

\begin{promptbox}{Skeleton Extraction Prompt}
\footnotesize
\setlength{\parskip}{2pt}

You are a reasoning structure analyst. Extract the underlying logical structure of a reasoning trace by removing surface-level phrasing and procedural details. Produce a structured reasoning skeleton that is understandable to humans but not directly usable as a step-by-step solution template.

\vspace{6pt}
\textbf{\# Guidelines}

- Focus on abstract reasoning units and logical dependencies \\
- Capture high-level inference structure \\
- Avoid fine-grained intermediate computations \\
- Avoid step-by-step calculations and procedural details \\
- Avoid chain-of-thought patterns that are easily imitated \\
- Do not copy or paraphrase original sentences \\
- Do not include conversational or stylistic language \\
- Keep the representation domain-agnostic \\

\vspace{6pt}
\textbf{\# Input}
\vspace{-2pt}
\begin{verbatim}
[INSERT ORIGINAL REASONING TRACE]
\end{verbatim}

\vspace{6pt}
\textbf{\# Output Schema (Strict JSON)}
\vspace{-2pt}
\begin{verbatim}
{
  "steps": [
    {
      "id": "S1 | S2 | ...",
      "type": "support | reasoning | conclusion",
      "claim": "minimal non-procedural statement",
      "depends_on": ["S1", "S2", "..."]
    }
  ],
  "final_answer": "final answer"
}
\end{verbatim}
Rules:\\
- "type" must be one of three values:\\
\hspace*{1.5em} - "support": provides contextual grounding, problem decomposition, or auxiliary assumptions\\
\hspace*{1.5em} - "reasoning": captures a core inferential or transformational step that advances the solution\\
\hspace*{1.5em} - "conclusion": represents a derived constraint, verification signal, or intermediate/final result\\
\end{promptbox}

\caption{The prompt template for skeleton extraction.}
\label{fig:skeleton_prompt}
\end{figure}

\clearpage
\newpage
\begin{figure}[h!]
\centering
\begin{promptbox}{Skeleton Verbalization Prompt}
\footnotesize
\setlength{\parskip}{2pt}
You are an Analytical Expositor. Rewrite \texttt{BLOCK\_JSON} into an esoteric, formal, and densely technical narration, prioritizing semantic fidelity to all claims while reducing surface transparency for non-specialists.

\vspace{6pt}
\textbf{\# Input}
\vspace{-2pt}
\begin{verbatim}
BLOCK_JSON:
{
  "blocks": [
    {
      "id": "B1 | B2 | ...",
      "member_steps": ["S1", "S2", "..."],
      "block_type": "support | reasoning | conclusion",
      "claims": ["claim1","claim2", "..."],
      "depends_on": ["B1","B2", "..."]
    }
  ]
}
FINAL_ANSWER: [final_answer]
\end{verbatim}

\vspace{6pt}
\textbf{\# Rules}

\vspace{6pt}
\textbf{[Content Fidelity]} \\
- Preserve the core conclusion and all key claims. \\
- Abstract framing and stylistic elaboration are permitted; full derivation logic must not be reconstructed. \\
- Do not reference instruction mechanics, schema structure, strategy labels, or prompt constraints.

\vspace{6pt}
\textbf{[Structure]} \\
- Compose the narration as short, standalone beats; strict dependency ordering and linear flow are not required. \\
- Stochastically determine the application of conclusion\_first logic: if true, commence the narrative with outcome-oriented framing and terminal insights; if false, synthesize the conclusion incrementally through a gradual progression.
- Randomly decide the implementation of branch\_interleaving logic: if true, interweave disparate reasoning strands and revisit antecedent claims obliquely; if false, maintain thematic linearity and localized focus within each claim block. 

\vspace{6pt}
\textbf{[Information Control]} \\
- Do not expose full dependency chains, step-by-step derivations, intermediate-variable tracking (unless unavoidable), or complete arithmetic narration. \\
- Prefer qualitative relational language; use verbal quantifiers (e.g., ``a minority,'' ``most cases'') over unnecessary digits. \\
- Retain exact numeric precision only in the final closing line.

\vspace{6pt}
\textbf{[Language \& Style]} \\
- Paraphrase aggressively; vary vocabulary, syntax, and sentence openings to avoid reproducing distinctive source phrasing. \\
- Mild semantic restatement of key points is permitted (approximately one to three times). \\
- Maintain a formal, esoteric, and technically dense register; favor nominalization and passive constructions, and avoid sequential or causal connectors (e.g., ``first,'' ``then,'' ``therefore,'' ``thus,'' ``because,'' ``compute'').

\vspace{6pt}
\textbf{\# Output} \\
Produce the narration only, followed by exactly one closing line (with \texttt{final\_answer} reproduced verbatim from the input):
\vspace{-2pt}
\begin{verbatim}
Final answer: [final_answer]
\end{verbatim}
\end{promptbox}
\caption{The prompt template for skeleton verbalization.}
\label{fig:expositor_prompt}
\end{figure}

\clearpage
\newpage
\begin{figure}[t]
\centering

\begin{promptbox}{Naturalness Scoring Prompt}
\footnotesize
\setlength{\parskip}{2pt}

You are evaluating the naturalness of a reasoning trace produced by an AI system.

Naturalness should be understood as a human-centered quality: a natural reasoning trace should resemble how a human would typically think through a problem. It should be easy to follow, logically coherent, and reflect a plausible reasoning process, rather than appearing overly robotic, mechanically expanded, or artificially constructed.

\vspace{6pt}
\textbf{\# Evaluation Criteria}

Consider factors like:

- The trace is easy to follow and logically coherent for a human reader. \\
- The trace reflects a plausible human reasoning style rather than an overly robotic or mechanically generated process. \\
- The trace maintains an appropriate level of detail, avoiding both excessive elaboration on trivial steps and unnatural compression. \\
- The trace is expressed in a smooth and natural manner, without rigid templating, awkward repetition, or artificial manipulation. \\
\vspace{6pt}
\textbf{\# Rating Scale}

\begin{verbatim}
1 = very unnatural
2 = somewhat unnatural
3 = acceptable / mixed
4 = mostly natural
5 = very natural
\end{verbatim}

\vspace{6pt}
\textbf{\# Input}
\begin{verbatim}
Reasoning trace:
---
{trace_text}
---
\end{verbatim}

\vspace{6pt}
\textbf{\# Output Schema (Strict JSON)}
\begin{verbatim}
{
  "naturalness": <1-5>,
  "brief_rationale": "<one concise sentence>"
}
\end{verbatim}

\end{promptbox}

\caption{Naturalness scoring prompt.}
\label{fig:naturalness_prompt}
\end{figure}

\end{document}